\documentclass[11pt]{article}
\usepackage{acl2015}
\usepackage{times}
\usepackage{url}
\usepackage{latexsym}
\usepackage{graphicx}
\usepackage{amsmath}
\usepackage{amsfonts}
\usepackage{pbox}

\title{Mixing Dirichlet Topic Models and Word Embeddings to Make lda2vec}

\author{Christopher Moody\\
  Stitch Fix One Montgomery Tower, Suite 1200 \\
  San Francisco, California 94104, USA\\
  \texttt{chrisemoody@gmail.com}
}

\date{}

\begin{document}
\maketitle
\begin{abstract}
Distributed dense word vectors have been shown to be
effective at capturing token-level semantic and syntactic regularities in language, 
while topic models can form interpretable representations over documents. 
In this work, we describe \textit{lda2vec}, a model that learns dense word vectors jointly with 
Dirichlet-distributed latent document-level mixtures of topic vectors.
In contrast to continuous dense document representations, this formulation
produces sparse, interpretable document mixtures through a non-negative simplex constraint.
Our method is simple to incorporate into existing automatic differentiation frameworks and allows
for unsupervised document representations geared for use by scientists while 
simultaneously learning word vectors and the linear relationships between them.
\end{abstract}

\section{Introduction}

\begin{figure}
  \includegraphics[trim=0cm 0cm 0cm 0cm, clip=true, width=\linewidth]{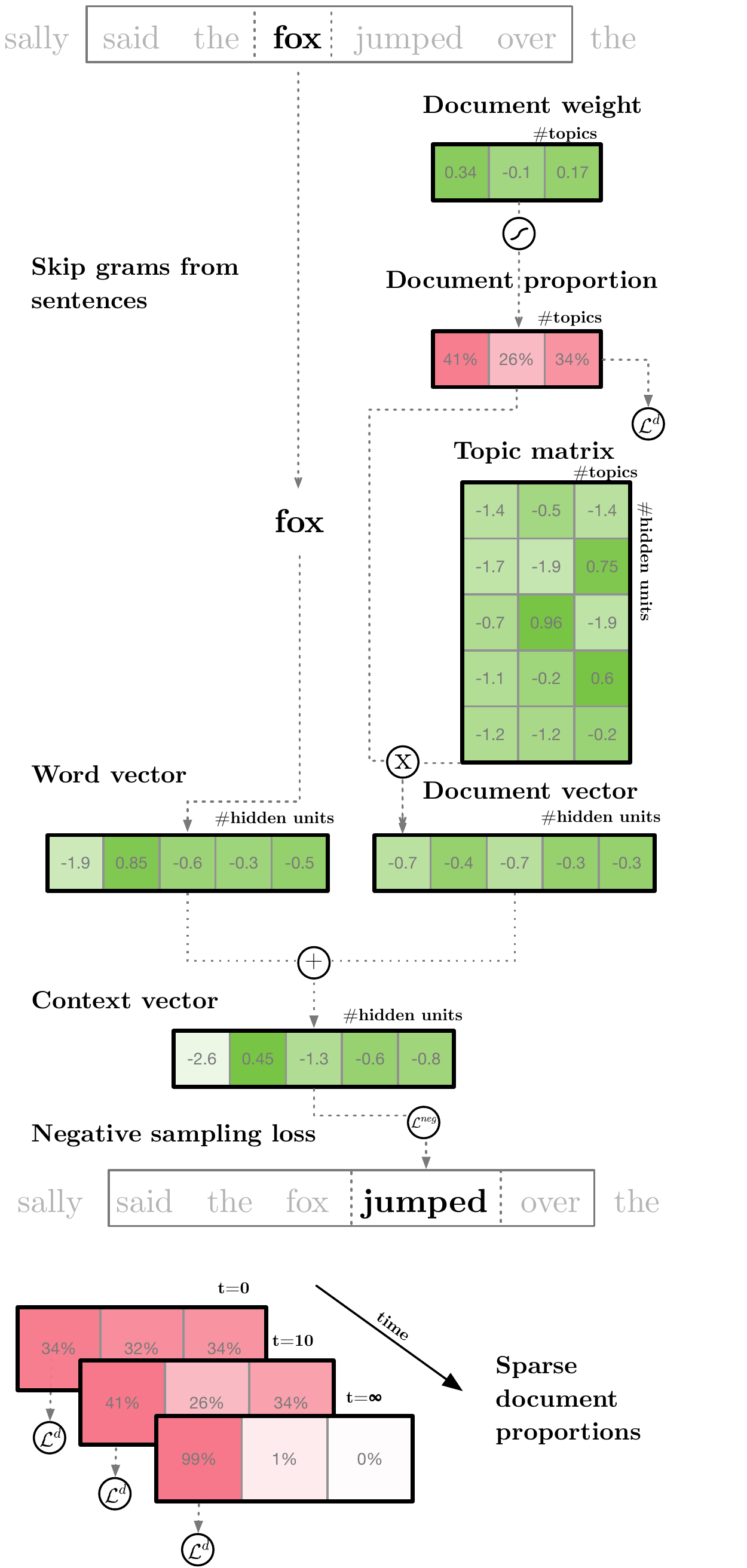}
  
  \caption{\textit{lda2vec} builds representations over both words and documents by mixing word2vec's skipgram architecture with Dirichlet-optimized sparse topic mixtures. The various components and transformations present in the diagram are described in the text.
  \label{fig:lda2vec_network}
  }
\end{figure}

Topic models are popular for their ability to organize document collections into a smaller set of prominent themes. 
In contrast to dense distributed representations, these document and topic representations 
are generally accessible to humans and more easily lend themselves to being interpreted. 
This interpretability provides additional options to highlight the patterns
and structures within our systems of documents. For example, 
using Latent Dirichlet Allocation (LDA) 
topic models can reveal cluster of words within documents \cite{Blei:2003tn}, 
highlight temporal trends \cite{Charlin:2015fy}, 
and infer networks of complementary products \cite{McAuley:2015kd}.
See \newcite{Blei:2010fu} for an overview of topic modelling 
in domains as diverse as computer vision, genetic markers, survey data,
and social network data.

Dense vector approaches to building document representations also exist:
\newcite{Le:2014vd} propose paragraph vectors that
are predictive of bags of words within paragraphs, \newcite{Kiros:2015uq} build
vectors that reconstruct the sentence sequences before and after a given sentence,
and \newcite{Ghosh:2016tia} construct contextual LSTMs that predict proceeding sentence
features. Probabilistic topic models tend to form
documents as a sparse mixed-membership of topics while neural network
models tend to model documents as dense vectors. 
By virtue of both their sparsity and low-dimensionality,
representations from the 
former are simpler to inspect and more immediately yield high level
intuitions about the underlying system
(although not without hazards, see \newcite{Chang:2009wd}). 
This paper explores
hybrid approaches mixing sparse document representations with dense 
word and topic vectors.

Unfortunately, crafting a new probabilistic topic model
requires deriving a new approximation, a procedure 
which takes substantial expertise and must be customized to every model. 
As a result, prototypes are time-consuming to develop
and changes to model architectures must be carefully considered.
However, with modern automatic differentiation frameworks
the practitioner can focus
development time on the model design rather than the model approximations.
This expedites the process of evaluating which model features are relevant.
This work takes advantage of the Chainer \cite{Tokui:tt} framework
to quickly develop models while also enabling us to utilize GPUs
to dramatically improve computational speed.

Finally, traditional topic models over text do not take advantage
of recent advances in distributed word representations which can capture
semantically meaningful regularities between tokens. 
The examination of word co-occurrences has proven to be a fruitful research paradigm.
For example, \newcite{Mikolov:2013uz} utilize Skipgram Negative-Sampling
(SGNS) to train word embeddings 
using word-context pairs formed from windows moving across a text corpus. 
These  vector representations ultimately encode remarkable linearities 
such as $king - man + woman = queen$.
In fact, \newcite{Levy:2014vd} demonstrate that this is implicitly
factorizing a variant of the Pointwise Mutual Information (PMI) matrix
that emphasizes predicting frequent co-occurrences over rare ones.
Closely related to the PMI matrix, \newcite{Pennington:2014jd} factorize
a large global word count co-occurrence matrix to yield more efficient
and slightly more performant computed embeddings than SGNS. 
Once created, these representations
are then useful for information retrieval \cite{Manning:2009kv} 
and parsing tasks \cite{Levy:2014gw}. In this work, we will
take advantage of word-level representations to build document-level
abstractions.

This paper extends distributed word representations 
by including interpretable document representations
and demonstrate that model inference can be performed and extended
within the framework of automatic differentiation.

\section{Model}

This section describes the model for \textit{lda2vec}. We are interested
in modifying the Skipgram Negative-Sampling (SGNS) objective in \cite{Mikolov:2013uz} to utilize
document-wide feature vectors while simultaneously learning continuous document
weights loading onto topic vectors. 
The network architecture is shown in Figure \ref{fig:lda2vec_network}. 

The total loss term $\mathcal{L}$ in \eqref{eq:objl} is the sum of the Skipgram Negative Sampling Loss
(SGNS) $\mathcal{L}_{ij}^{neg}$ with the addition of a Dirichlet-likelihood term over document weights,  $\mathcal{L}^d$ that will be discussed later. The loss is conducted using a context vector, $\vec{c_j}$, pivot word vector $\vec{w_j}$, target word vector $\vec{w_i}$, and negatively-sampled word vector $\vec{w_l}$.

\begin{align} 
\label{eq:objl} 
\mathcal{L} &= \mathcal{L}^{d} + \Sigma_{ij} \mathcal{L}_{ij}^{neg} \\
\label{eq:objns} \mathcal{L}_{ij}^{neg} &= \log \sigma(\vec{c_j} \cdot \vec{w_i}) + \Sigma_{l=0}^{n}\log\sigma(-\vec{c_j}\cdot \vec{w_{l}})
\end{align}
 
 \subsection{Word Representation}
 
 As in \newcite{Mikolov:2013uz}, pairs of
 pivot and target words $(j, i)$ are extracted when they co-occur in a moving window scanning across the corpus.
In our experiments, the window contains five tokens before and after the pivot token.
For every pivot-target pair of words the pivot word is used to predict the nearby target word.
Each word is represented with a fixed-length dense distributed-representation vector, 
but unlike \newcite{Mikolov:2013uz} the same word vectors are used in both the pivot and target representations. 
The SGNS loss shown in \eqref{eq:objns} attempts to discriminate context-word pairs
that appear in the corpus from those randomly sampled from a `negative' pool of
words. 
This loss is minimized when the observed words are completely separated from 
the marginal distribution. 
The distribution from which tokens are drawn is $u^{\beta}$, where $u$
denotes the overall word frequency normalized by the total corpus size.
Unless stated otherwise, the negative sampling power $beta$ is set to $3/4$ and the number of negative samples is fixed to $n=15$ as in \newcite{Mikolov:2013uz}. 
Note that a distribution of $u^{0.0}$ would draw negative tokens
from the vocabulary with no notion of popularity while a distribution
proportional with $u^{1.0}$ draws from the empirical unigram distribution. 
Compared to the unigram distribution, 
the choice of $u^{3/4}$ slightly emphasizes choosing infrequent words for negative samples.
In contrast to optimizing the softmax cross entropy, which requires modelling
the overall popularity of each token, 
negative sampling focuses on learning word vectors conditional on a context
by drawing negative samples from each token's marginal popularity in the corpus.

\subsection{Document Representations}

\textit{lda2vec} embeds both words and document vectors into the same space and trains
both representations simultaneously. By adding the pivot and document vectors together,
both spaces are effectively joined.
\newcite{Mikolov:2013uz} provide the intuition that word vectors can be summed together
to form a semantically meaningful combination of both words. For example, the vector representation
 for $Germany + airline$ is similar to  the vector for $Lufthansa$. We would like to exploit the
 additive property of word vectors to construct a meaningful sum of word and document vectors.
For example, if as \textit{lda2vec} is scanning a document the $j$th  word is $Germany$,
then neighboring words are predicted to be similar such as $France$, $Spain$, and $Austria$.
But if the document is specifically about airlines, then we would like to construct a document vector similar to the word vector for $airline$. Then instead of predicting tokens similar to $Germany$ alone, 
predictions similar to both the document and the pivot word can be made such as: $Lufthansa$, 
$Condor\ Flugdienst$, and $Aero\ Lloyd$. Motivated by the meaningful sums of words vectors, 
in \textit{lda2vec} the context vector is explicitly designed to be the sum of a document vector
and a word vector as in \eqref{eq:objcj}:

\begin{align} 
\label{eq:objcj}\vec{c_j} &= \vec{w_j} + \vec{d_j}
\end{align}

This models document-wide relationships by preserving $\vec{d_j}$ for all word-context pairs
in a document, while still leveraging local inter-word relationships stemming from 
the interaction between the pivot word vector $\vec{w_j}$ and target word $\vec{w_i}$.
 The document and word vectors are summed together to form a
context vector that intuitively captures long- and short-term themes, respectively. 
In order to prevent co-adaptation, we also perform dropout on both the 
unnormalized document vector $\vec{d_j}$ and the pivot word vector $\vec{w_j}$ 
\cite{Hinton:2012tv}.

\subsubsection{Document Mixtures}

If we only included structure up to this point, the model would produce a 
dense vector for every document.
However, \textit{lda2vec} strives to form interpretable representations
and to do so an additional constraint is imposed such that the document representations
are similar to those in traditional LDA models. 
We aim to generate a document vector from a mixture of topic vectors
and to do so, we begin by constraining the document vector $\vec{d_j}$ 
to project onto a set of latent topic vectors 
${\vec{t_0}, \vec{t_1}, ...,\vec{t_k}}$:

\begin{equation} 
\label{eq:objdj}\vec{d_j} = p_{j0} \cdot \vec{t_0} + p_{j1} \cdot \vec{t_1}
+ ... + p_{jk} \cdot \vec{t_k}
+ ... + p_{jn} \cdot \vec{t_n}
\end{equation}

Each weight $0\leq p_{jk}\leq1$ is a fraction that denotes the membership of document $j$ in the
topic $k$. 
For example, the Twenty Newsgroups model described later has 11313 documents
and  $n=20$ topics so $j=0...11312$, $k=0...19$. 
When the word vector dimension is set to 300, it is assumed that 
the document vectors $\vec{d_j}$,  
word vectors $\vec{w_i}$ and topic vectors $\vec{t_k}$ all have dimensionality 300.
Note that the topics $\vec{t_k}$ are shared and are a
common component to all documents but whose strengths are modulated by
document weights $p_{jk}$ that are unique to each document. 
To aid interpretability, the document memberships are designed to be non-negative,
and to sum to unity. 
To achieve this constraint, a softmax transform maps latent vectors initialized in 
$\mathbb{R}^{300}$ onto the simplex defined by $p_{jk}$. 
The softmax transform naturally enforces the constraint that $\Sigma_k p_{jk}=1$
and allows us interpret memberships as percentages rather than unbounded weights.

Formulating the mixture in \eqref{eq:objdj} as a sum 
ensures that topic vectors $\vec{t_k}$, document vectors $\vec{d_j}$ and 
word vectors $\vec{w_i}$, operate in the same space.
As a result, what words $\vec{w_i}$ are most similar to
any given topic vector $\vec{t_k}$ can be directly calculated.
While each topic is not literally a token present in the corpus, 
it's similarity to other tokens is meaningful and can be measured. 
Furthermore, by examining the list of most similar words
one can attempt to interpret what the topic represents.
For example, by calculating the most similar token to any
topic vector (e.g. $argmax_i(\vec{t_0} \cdot \vec{w_i})$) one may discover that
the first topic vector $\vec{t_{0}}$ is similar to the tokens
\textit{pitching}, \textit{catcher}, and \textit{Braves} while the second topic vector $\vec{t_{1}}$
may be similar to \textit{Jesus}, \textit{God}, and \textit{faith}. This provides us the
option to interpret the first topic as \textit{baseball} topic, and as a result 
the first component in every document proportion $p_{j0}$ indicates how much 
document $j$ is in the \textit{baseball} topic. 
Similarly, the second topic may be interpreted
as \textit{Christianity} and the second component of any document proportion $p_{j1}$
indicates the membership of that document in the \textit{Christianity} topic.

\subsubsection{Sparse Memberships}
Finally, the document weights $p_{ij}$ are sparsified by optimizing the document
weights with respect to a Dirichlet likelihood with a low concentration parameter
$\alpha$:

\begin{align} 
\label{eq:objdl} \mathcal{L}^{d} &=\lambda \Sigma_{jk}\ (\alpha - 1) \log p_{jk}
\end{align}

The overall objective in \eqref{eq:objdl} measures the likelihood of document $j$
in topic $k$ summed over all available documents. 
The strength of this term is modulated by the tuning parameter $\lambda$.
This simple likelihood encourages the document proportions coupling in each
topic to be sparse when $\alpha<1$ and homogeneous when $\alpha>1$.
To drive interpretability, we are interested in finding sparse memberships and so set 
$\alpha={n}^{-1}$ where $n$ is the number of topics.  
We also find that setting the overall strength of the Dirichlet 
optimization to $\lambda=200$ works well.
Document proportions are initialized to be relatively homogeneous, 
but as time progresses, the $\mathcal{L}^d$ 
encourages document proportions vectors to become 
more concentrated (e.g. sparser) over time. 
In experiments without this
sparsity-inducing term (or equivalently when $\alpha=1$) 
the document weights $p_{ij}$ tend to have probability 
mass spread out among all elements. Without any sparsity 
inducing terms the existence of so many non-zero weights makes
interpreting the document vectors difficult. Furthermore, we find that
the topic basis are also strongly affected, and the topics become incoherent.

\subsection{Preprocessing and Training}
 The objective in \eqref{eq:objl} is trained in individual minibatches at a time while using the Adam optimizer \cite{Kingma:2014us} for two hundred epochs across the dataset. The Dirichlet likelihood term $\mathcal{L}^d$ is typically computed over all documents, so in modifying the objective to minibatches we adjust the loss of the term to be proportional to the minibatch size divided by the size of the total corpus.
 Our software is open source, available online, documented and unit tested\footnote{The code for \textit{lda2vec} is available online at \url{https://github.com/cemoody/lda2vec}}. 
 Finally, the top ten  most likely words in a given topic are submitted to the online \textit{Palmetto}\footnote{The online evaluation tool can be accessed at \url{http://palmetto.aksw.org/palmetto-webapp/}} topic quality measuring tool and the coherence measure $C_v$ is recorded. After evaluating multiple alternatives, 
 $C_v$ is the recommended coherence metric in \newcite{Roder:2015ev}.
 This measure averages the Normalized Pointwise Mutual Information (NPMI) for every pair of words within a sliding window of size 110 on an external corpus and returns mean of the NPMI for the submitted set of words. Token-to-word similarity is evaluated using the \texttt{3COSMUL} measure \cite{Levy:2014wb}.

\begin{figure}
\begin{tabular}{|r|r|r|}
\hline
 \# of topics &  $\beta$ &  Topic Coherences \\
 \hline
       20 &  0.75 &    \textbf{0.567} \\
       30 &  0.75 &    0.555 \\
       40 &  0.75 &    0.553 \\
       50 &  0.75 &    0.547 \\
       20 &  1.00 &    0.563 \\
       30 &  1.00 &    0.564 \\
       40 &  1.00 &    0.552 \\
       50 &  1.00 &    0.558 \\
\hline
\end{tabular}

\caption{Average topic coherences found by \textit{lda2vec} in the Twenty Newsgroups dataset are given. The topic coherence has been demonstrated to correlate with human evaluations of topic models \cite{Roder:2015ev}. The number of topics chosen is given, as well as the negative sampling exponent parameter $\beta$.  Compared to $\beta=1.00$, $\beta=0.75$ draws more rare words as negative samples. The best topic coherences are found in models $n=20$ topics and a $\beta=0.75$. \label{fig:20ngcoherence} }

\end{figure}

\begin{figure*}
\centering
\begin{tabular}{|l|l|l|l|l|}
\hline
Topic Label  &      ``Space'' &     ``Encryption'' &    ``X Windows'' &   ``Middle East'' \\
\hline
Top tokens &  astronomical &  encryption &  mydisplay &  Armenian \\
 &     Astronomy &     wiretap &       xlib &  Lebanese \\
 &    satellite &     encrypt &     window &    Muslim \\
 &    planetary &      escrow &     cursor &      Turk \\
 &    telescope &     Clipper &     pixmap &        sy \\
\hline
Topic Coherence &        0.712 &       0.675 &      0.472 &     0.615 \\
\hline
\end{tabular}
\caption{Topics discovered by \textit{lda2vec} in the Twenty Newsgroups dataset. The inferred
topic label is shown in the first row. The tokens with highest similarity to the topic are shown
immediately below. Note that the twenty newsgroups corpus contains corresponding 
newsgroups such as  \textit{sci.space}, \textit{sci.crypt},
\textit{comp.windows.x} and  \textit{talk.politics.mideast}. \label{fig:20ngtopics}}

\end{figure*}

\section{Experiments}

\subsection{Twenty Newsgroups}

This section details experiments in discovering the salient topics in the Twenty Newsgroups dataset, a popular corpus for machine learning on text. 
Each document in the corpus was posted to one of twenty possible newsgroups. 
While the text of each post is available to \textit{lda2vec},
each of the newsgroup partitions is not revealed to the algorithm
but is nevertheless useful for post-hoc qualitative evaluations of the discovered topics.
The corpus is preprocessed using the data loader 
available in Scikit-learn \cite{Pedregosa:2012tv} and tokens are identified using the SpaCy parser \cite{Honnibal:2015jm}. Words are lemmatized to group multiple inflections into single tokens. Tokens that occur fewer than ten times in the corpus are removed, as are tokens that appear to be URLs, numbers or contain special symbols within their orthographic forms. After preprocessing, the dataset contains 1.8 million observations of 8,946 unique tokens in 11,313 documents. Word vectors are initialized to the pretrained values found in \newcite{Mikolov:2013uz} but otherwise updates are allowed to these vectors at training time.

A range of \textit{lda2vec} parameters are evaluated by varying the number of topics $n\in{20, 30, 40, 50}$ and the negative sampling exponent $\beta\in{0.75, 1.0}$. The best topic coherences were achieved with $n=20$ topics and with negative sampling power $\beta=0.75$ as summarized in Figure \ref{fig:20ngcoherence}. We briefly experimented with variations on dropout ratios but we did not observe any substantial differences.

Figure \ref{fig:20ngtopics} lists four example topics discovered in the Twenty Newsgroups dataset. Each topic is associated with a topic vector that lives in the same space as the trained word vectors and listed are the most similar words to each topic vector. 
The first topic shown has high similarity to the tokens \textit{astronomical}, \textit{Astronomy}, \textit{satellite}, \textit{planetary}, and \textit{telescope} and is thus likely a `Space'-related topic similar to the  `sci.space' newsgroup. 
The second example topic is similar to words semantically related to `Encryption', such as \textit{Clipper} and \textit{encrypt}, and is likely related to the `sci.crypt' newsgroup. The third and four example topics are `X Windows' and `Middle East' which likely belong to the `comp.windows.x' and `talk.politics.mideast' newsgroups.

\begin{figure*}
\centering

\begin{tabular}{|l|l|l|l|l|l|}
\hline
  ``Housing Issues'' &  ``Internet Portals'' &   ``Bitcoin'' &  ``Compensation'' &  ``Gadget Hardware'' \\
\hline
        more housing &     DDG. &       btc &    current salary &  the Surface Pro \\
        basic income &     Bing &  bitcoins &       more equity &             HDMI \\
         new housing &  Google+ &   Mt. Gox &           vesting &   glossy screens \\
        house prices &      DDG &     MtGox &            equity &          Mac Pro \\
  short-term rentals &  iGoogle &       Gox &  vesting schedule &      Thunderbolt \\
\hline
\end{tabular}

\caption{Topics discovered by \textit{lda2vec} in the Hacker News comments dataset. The inferred
topic label is shown in the first row. We form tokens from noun phrases  to capture the unique vocabulary of this specialized corpus. \label{fig:hntopics} }
\end{figure*}

\begin{figure*}
\begin{tabular}{|l|l|l|l|l|l}
\hline
Artificial sweeteners &        Black holes &       Comic Sans & Functional Programming &  San Francisco \\
\hline
               glucose &          particles &         typeface &                     FP &       New York \\
              fructose &      consciousness &            Arial &                Haskell &      Palo Alto \\
                  HFCS &           galaxies &        Helvetica &                    OOP &            NYC \\
                sugars &  quantum mechanics &  Times New Roman &   functional languages &  New York City \\
                 sugar &           universe &             font &                 monads &             SF \\
               Soylent &        dark matter &         new logo &                   Lisp &  Mountain View \\
            paleo diet &           Big Bang &    Anonymous Pro &                Clojure &        Seattle \\
                  diet &            planets &      Baskerville &        category theory &    Los Angeles \\
         carbohydrates &       entanglement &       serif font &                     OO &         Boston \\
\hline

\end{tabular}
\caption{Given an example token in the top row, the most similar words available in the Hacker News comments corpus are reported. \label{fig:hnsim} }
\end{figure*}

\subsection{Hacker News Comments corpus}

This section evaluates \textit{lda2vec} on a very large corpus of Hacker News \footnote{See \url{https://news.ycombinator.com/}} comments.  Hacker News is social content-voting website and community whose focus is largely on technology and entrepreneurship. In this corpus, a single document is composed of all of the words in all comments posted to a single article. 
Only stories with more than 10 comments are included, and only comments from users with more than 10 comments are included. 
We ignore other metadata such as votes, timestamps, and author identities. 
The raw dataset \footnote{The raw dataset is freely available at \url{https://zenodo.org/record/45901}} is available for download online. The corpus is nearly fifty times the size of the Twenty Newsgroups corpus which is sufficient for learning a specialized vocabulary. 
To take advantage of this rich corpus, we use the SpaCy to tokenize whole noun phrases and entities at once \cite{Honnibal:2015jm}. 
The specific tokenization procedure\footnote{The tokenization procedure is available online at \url{https://github.com/cemoody/lda2vec/blob/master/lda2vec/preprocess.py}} is also available online, as are the preprocessed datasets \footnote{A tokenized dataset is freely available at \url{https://zenodo.org/record/49899}}  results. 
This allows us to capture phrases such as \textit{community policing measure} and prominent figures such as \textit{Steve Jobs} as single tokens. 
However, this tokenization procedure generates a vocabulary substantially different from the one available in the Palmetto topic coherence tool and so we do not report topic coherences on this corpus. After preprocessing, the corpus contains 75 million tokens in 66 thousand documents with 110 thousand unique tokens. Unlike the Twenty Newsgroups analysis, word vectors are initialized randomly instead of using a library of pretrained vectors.

We train an \textit{lda2vec} model using 40 topics and 256 hidden units and report the learned topics that demonstrate the themes present in the corpus. Furthermore, we demonstrate that word vectors and semantic relationships specific to this corpus are learned.

In Figure \ref{fig:hntopics} five example topics discovered by \textit{lda2vec} in the Hacker News corpus are listed. These topics demonstrate that the major themes of the corpus are reproduced and represented in learned topic vectors in a similar fashion as in LDA \cite{Blei:2003tn}.
The first, which we hand-label \textit{Housing Issues} has prominent tokens relating to housing policy issues such as housing supply (e.g. \textit{more housing}), and costs (e.g. \textit{basic income} and \textit{house prices}). Another topic lists major internet portals, such as the privacy-conscious search engine `Duck Duck Go' (in the corpus abbreviated as \textit{DDG}), as well as other major search engines (e.g. \textit{Bing}), and home pages (e.g. \textit{Google+}, and \textit{iGoogle}). A third topic is that of the popular online curency and payment system \textit{Bitcoin}, the abbreviated form of the currency \textit{btc}, and the now-defunct Bitcoin trading platform \textit{Mt. Gox}. A fourth topic considers salaries and compensation with tokens such as \textit{current salary}, \textit{more equity} and \textit{vesting}, the process by which employees secure stock from their employers. A fifth example topic is that of technological hardware like \textit{HDMI} and \textit{glossy screens} and includes devices such as \textit{the Surface Pro} and \textit{Mac Pro}.

Figure \ref{fig:hnsim} demonstrates that token similarities are learned in a similar fashion as in SGNS \cite{Mikolov:2013uz} but specialized to the Hacker News corpus. Tokens similar to the token \textit{Artificial sweeteners} include other sugar-related tokens like \textit{fructose} and food-related tokens such as \textit{paleo diet}. Tokens similar to \textit{Black holes} include physics-related concepts such as \textit{galaxies} and \textit{dark matter}. 
The Hacker News corpus devotes a substantial quantity of text to fonts and design, and the words most similar to \textit{Comic Sans} are other popular fonts  (e.g. \textit{Times New Roman} and \textit{Helvetica}) as well as font-related concepts such as \textit{typeface} and \textit{serif font}. Tokens similar to \textit{Functional Programming} demonstrate similarity to other computer science-related tokens while tokens similar to \textit{San Francisco} include other large American cities as well smaller cities located in the San Francisco Bay Area.

\begin{figure}

\begin{tabular}{|p{4cm}|l|}
\hline
Query &                Result \\
\hline
California + technology             &   Silicon Valley \\
\hline
digital + currency                  &          Bitcoin \\
\hline
Javascript - browser + server       &          Node.js \\
\hline
Mark Zuckerberg - \newline Facebook + Amazon &       Jeff Bezos \\
\hline
NLP - text + image                  &  computer vision \\
\hline
Snowden - United States + Sweden    &          Assange \\
\hline
Surface Pro - Microsoft + Amazon    &           Kindle \\
\hline

\end{tabular}

\caption{Example linear relationships discovered by \textit{lda2vec} in the Hacker News comments dataset. The first column indicates the example input query, and the second column indicates the token most similar to the input. \label{fig:hnrel}}

\end{figure}

Figure \ref{fig:hnrel} demonstrates that in addition to learning topics over documents and similarities to word tokens, 
linear regularities between tokens are also learned. The `Query' column lists a selection of tokens that when combined yield a token vector closest to the token shown in the `Result' column. The subtractions and additions of vectors are evaluated literally, but instead take advantage of the \texttt{3COSMUL} objective \cite{Levy:2014wb}. The results show that relationships between tokens important to the Hacker News community exists between the token vectors. For example, the vector for \textit{Silicon Valley} is similar to both \textit{California} and \textit{technology}, 
\textit{Bitcoin} is indeed a \textit{digital currency}, 
\textit{Node.js} is a technology that enables running \textit{Javascript} on \textit{servers} instead of on client-side \textit{browsers},
\textit{Jeff Bezos} and \textit{Mark Zuckerberg} are CEOs of \textit{Amazon} and \textit{Facebook} respectively,
\textit{NLP} and \textit{computer vision} are fields of machine learning research primarily dealing with \textit{text} and \textit{images} respectively,
Edward \textit{Snowden} and Julian \textit{Assange} are both whistleblowers who were primarily located in the \textit{United States} and \textit{Sweden}
and finally the \textit{Kindle} and the \textit{Surface Pro} are both tablets manufactured by \textit{Amazon} and \textit{Microsoft} respectively. In the above examples semantic relationships between tokens encode for attributes and features including: location, currencies, server v.s. client, leadership figures, machine learning fields, political figures, nationalities, companies and hardware.

\subsection{Conclusion}

This work demonstrates a simple model, \textit{lda2vec}, that extends SGNS \cite{Mikolov:2013uz} to build unsupervised document representations that yield coherent topics. 
Word, topic, and document vectors are jointly trained and embedded in a common representation space that preserves semantic regularities between the learned word vectors while still yielding sparse and interpretable document-to-topic proportions in the style of LDA \cite{Blei:2003tn}. Topics formed in the Twenty Newsgroups corpus yield high mean topic coherences which have been shown to correlate with human evaluations of topics \cite{Roder:2015ev}. When applied to a Hacker News comments corpus, \textit{lda2vec} discovers the salient topics within this community and learns linear relationships between words that allow it solve word analogies in the specialized vocabulary of this corpus.
Finally, we note that our method is simple to implement in automatic differentiation frameworks and can lead to more readily interpretable unsupervised representations.

\bibliographystyle{acl}
\bibliography{bib2016}

\end{document}